\documentclass[12pt, number, 1p]{elsarticle}
\usepackage{lineno}
\modulolinenumbers[1]
\usepackage[utf8]{inputenc} %
\usepackage[T1]{fontenc}    %
\usepackage{hyperref}       %
\usepackage{url}            %
\usepackage{booktabs}       %
\usepackage[normalem]{ulem}
\useunder{\uline}{\ul}{}
\usepackage{amsfonts}       %
\usepackage{amsmath}
\usepackage{nicefrac}       %
\usepackage{microtype}      %
\usepackage{lipsum}		%
\usepackage{graphicx}
\usepackage[numbers]{natbib}
\usepackage{doi}
\usepackage{xcolor}
\usepackage{marginnote}
\usepackage{multirow}
\usepackage{tablefootnote}
\usepackage[justification=centering]{caption}
\usepackage{subcaption}
\usepackage{parskip}
\usepackage[backgroundcolor=black!20,linecolor=black!20,textsize=tiny]{todonotes}
\usepackage{placeins}
\usepackage{adjustbox}

\biboptions{authoryear}

\begin{document}
    \begin{frontmatter}
    
    \title{Lameness detection in dairy cows using pose estimation and bidirectional LSTMs}

    \author[1]{Helena Russello
	\corref{correspondingauthor1}
    }
	\cortext[correspondingauthor1]{Corresponding authors}
	\ead{firstname.lastname@wur.nl}
    
    \author[1]{Rik van der Tol}
    \author[1]{Eldert J. van Henten}
    \author[1]{Gert Kootstra
	\corref{correspondingauthor1}
    }
    \ead{firstname.lastname@wur.nl}
 	
    \address[1]{Agricultural Biosystems Engineering group, Wageningen University \& Research, Wageningen, The Netherlands }

    \begin{abstract}

This study presents a lameness detection approach that combines pose estimation and Bidirectional Long-Short-Term Memory (BLSTM) neural networks.
Combining pose-estimation and BLSTMs classifier offers the following advantages: markerless pose-estimation, elimination of manual feature engineering by learning temporal motion features from the keypoint trajectories, and working with short sequences and small training datasets. 
Motion sequences of nine keypoints (located on the cows' hooves, head and back) were extracted from videos of walking cows with the T-LEAP pose estimation model.
The trajectories of the keypoints were then used as an input to a BLSTM classifier that was trained to perform binary lameness classification. 
Our method 
significantly outperformed an established method that relied on manually-designed locomotion features: our best architecture achieved a classification accuracy of 85\%, against 80\% accuracy for the feature-based approach.
Furthermore, we showed that our BLSTM classifier could detect lameness with as little as one second of video data.

\end{abstract}


    \end{frontmatter}

\section{Introduction}

Lameness is a prevalent condition in dairy cows, and is characterized by an abnormal gait due to lesions in their hooves or limbs. 
Lameness has negative welfare and economic impacts as it affects milk production, fertility, and life quality of the cows~\citep{enting1997economic, whay2017impact}.
Lameness is commonly detected during visual locomotion scoring sessions where trained observers assess the lameness prevalence of the herd,  but these are time-consuming and performed sporadically~\citep{o2020invited}.
Automating locomotion scoring, e.g. by means of continuous camera monitoring, could allow for earlier lameness detection, and thereby timely treatment.

Video cameras are attractive sensors for automatic locomotion scoring because they are relatively inexpensive, non-intrusive, and scale well with large herds. 
Automatic lameness detection from videos is commonly approached by using locomotion traits from clinical gait scoring methods~\citep{sprecher_lameness_1997} to classify the degree of lameness in cows. 
For example, 
studies by~\cite{poursaberi_real-time_2010, viazzi_comparison_2014, van_hertem_implementation_2018} used the posture of the back,~\cite{song_automatic_2008} used the tracking distance,~\cite{wu_lameness_2020, zheng_cows_2023} used the step size, and~\cite{kang_accurate_2020} used the supporting phase to classify lameness.
Other studies, such as
~\cite{zhao_automatic_2023, barney2023deep, taghavi2024keeping, russello2024video} combined multiple locomotion traits as input features for lameness classification. 
Combining the back posture, head position, tracking distance, and stride length led to an improved classification performance over using single features~\citep{russello2024video}.
Although models using locomotion traits have shown promising results in detecting lameness, they have some limitations.
For instance, these locomotion traits assume low-noise data and may perform poorly with noisy data~\cite{taghavi2023cow}.
Additionally, manually selecting locomotion traits as features can restrict the information provided to such models and it may fail to capture complex patterns using only hand-crafted features. 

In order to address these limitations, several studies have used deep neural networks to perform both feature extraction and lameness classification.
Most of these methods consist of two to three steps: first isolating the body structure of the animal from video-frames, then performing feature extraction and lameness classification (the last two are often combined as one step).
For instance,~\cite{karoui_deep_2021} tracked adhesive physical markers placed on the legs of cows, and trained a Convolutional Neural Network (CNN) to predict lameness from the motion of said markers. 
~\cite{wu_lameness_2020} extracted leg coordinates with YOLOv3~\citep{farhadi2018yolov3} to create a step-size feature. 
Sequences of step-size vectors were then used in a Long-Short-Term Memory neural network (LSTM) and in a Bi-directional LSTM (BLSTM) to detect lameness.
~\cite{arazo_segmentation_2022} applied a segmentation model alongside the SlowFast video-recognition model~\citep{feichtenhofer2019slowfast} to classify lameness based on time-series of the body contour of the cows.
~\cite{jiang_single-stream_2020} extracted sequences of optical-flow maps, which captured the motion between video-frames, and used it in a BLSTM network to classify lameness.
Both~\cite{arazo_segmentation_2022} and~\cite{jiang_single-stream_2020} demonstrated that their models could classify lameness effectively using RGB video data alone (a one-step approach). 
However, they found that the classification accuracy was improved when using a two-step approach that incorporated segmentation masks~\citep{arazo_segmentation_2022} or optical-flow maps~\citep{jiang_single-stream_2020} to guide the model’s focus on the body structure of the animal.

Even though the use of deep learning models reduces the need for extensive pre-processing and manual feature engineering, while increasing robustness, the multi-steps methods discussed in the previous paragraph still have inherent limitations.
Leg-only tracking~\citep{wu_lameness_2020, karoui_deep_2021} potentially overlooks valuable motion patterns from other body parts,
whereas whole-body analysis~\citep{jiang_single-stream_2020, arazo_segmentation_2022} lacks specificity in tracking critical anatomical features, and its high dimensionality typically requires more training data, which makes the learning process more challenging.

To overcome these constraints, we propose a two-step approach that combines the advantages of pose estimation and BLSTM models~\citep{graves2005framewise}. %
We use the T-LEAP pose estimation model~\citep{russello_t-leap_2021}, which allows tracking of multiple keypoints, including leg joints, head position, and spine movement, thereby providing the classification model with compact yet comprehensive motion data from bio-mechanically relevant keypoints. %
Building upon the work of~\cite{russello2024video}, which demonstrated the effectiveness of keypoint trajectories for computing hand-crafted locomotion features, we use BLSTM models to directly learn locomotion features from sequences of keypoint trajectories and classify lameness. 
A BLSTM is a type of neural network that excels at capturing long-term relationships in sequential data and modeling complex temporal patterns. BLSTMs have been successfully used in diverse application domains, including human motion analysis~\citep{du2019bio, battistone_tglstm_2019}, and also lameness detection~\citep{jiang_single-stream_2020, wu_lameness_2020}.
In these application domains~\citep{du2019bio, jiang_single-stream_2020, wu_lameness_2020}, BLSTMs have consistently outperformed regular (unidirectional) LSTMs.
Therefore, we used BLSTMs for detecting lameness based on the keypoint trajectories of walking cows.

In summary, our proposed method offers the following advantages: the pose estimation provides a compact data representation of the spatio-temporal motion from multiple body parts, while the BLSTM model eliminates the need for noise filtering as well as manual feature engineering.
Our contributions are as follows:
\begin{enumerate}
    \item We present a two-step approach using BLSTM neural networks for detecting lameness from time-series data of nine keypoints. 
    Our research explores whether detecting lameness through learned locomotion features outperforms previously established manually-designed locomotion traits. 
    To ensure a fair assessment, we conduct a direct comparison with~\cite{russello2024video} using the same dataset.
    \item We evaluate the performance of our approach across multiple BLSTM architectures (different number of layers and layer sizes). Additionally, we evaluate the performance of the BLSTM classifiers across different sequence lengths (30, 60, and 90 frames, corresponding to 1, 2, and 3 seconds of video at 30 FPS), since healthy and lame cows tend to walk at different speeds~\citep{flower2005hoof}. 
    \item To promote reproducibility and facilitate future research, we will make both our data and code publicly available upon publication.

\end{enumerate}

\section{Materials and Methods}

\begin{figure}[ht]
    \centering
    \includegraphics[width=1\linewidth]{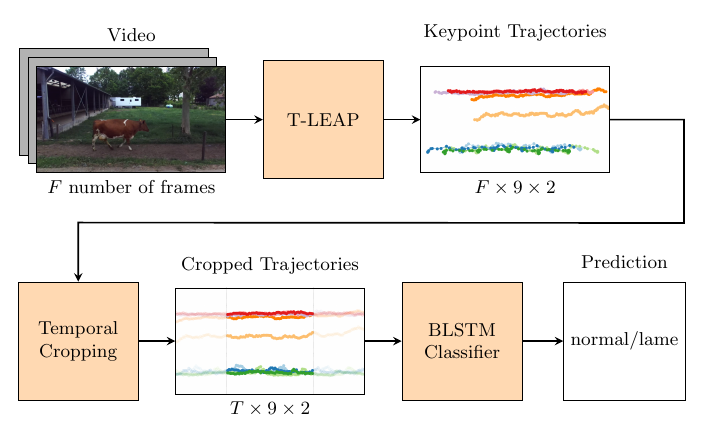}
    \caption{Outline of our lameness detection approach. The keypoint trajectories are extracted from videos using T-LEAP (Fig.~\ref{fig:trajectory-example-normal} and ~\ref{fig:trajectory-example-lame}). The keypoint trajectories are then trimmed to a fixed sequence length $T$ and passed to the BLSTM classifier.}
    \label{fig:lstm-workflow}
\end{figure}

Our approach, illustrated in Figure~\ref{fig:lstm-workflow}, uses T-LEAP to extract keypoint trajectories from videos of walking cows.
The keypoint trajectories are cropped to a fixed sequence length, and passed to a BLSTM classification model. 
The data and approach are further described in the following subsections.

\subsection{Dataset}
The experiments used keypoint trajectories from the dataset introduced in~\cite{russello2024video}, comprising 272 videos of 98 individual Holstein-Friesian cows. 
A ZED camera\footnote{\url{https://www.stereolabs.com/zed}} was used to film the cows from the side as they walked freely through an outdoor walkway.
The videos were recorded at 30 frames per second, with lengths ranging from 90 to 207 frames (mean length: 134 frames).

The dataset from~\cite{russello2024video} consists of keypoint trajectories extracted from the 272 videos.
For each video, four observers scored the gait using the~\cite{sprecher_lameness_1997} scale.
The gait scores were merged into binary lameness labels (\textit{normal}/\textit{lame}). 
Out of the 272 keypoint trajectories, 143 were labeled as \textit{normal}, and 129 were labeled as \textit{lame}, yielding a relatively balanced dataset.

\begin{figure}[ht]
    \centering
    \includegraphics[]{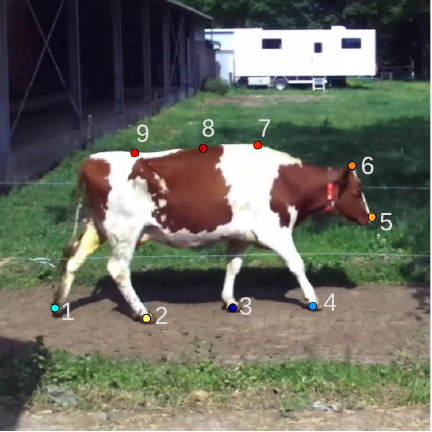}
    \caption{The 9 keypoints as described in~\cite{russello2024video}. The keypoints are named as follows: 1: Left-hind hoof, 2: Right-hind hoof, 3: Left-front hoof, 4: Right-front hoof 5: Nose, 6: Forehead, 7: Withers, 8: Caudal thoracic vertebrae, 9: Sacrum.}
    \label{fig:9keypoints}
\end{figure}

\begin{figure}[ht]
    \centering
    \includegraphics[width=0.8\textwidth]{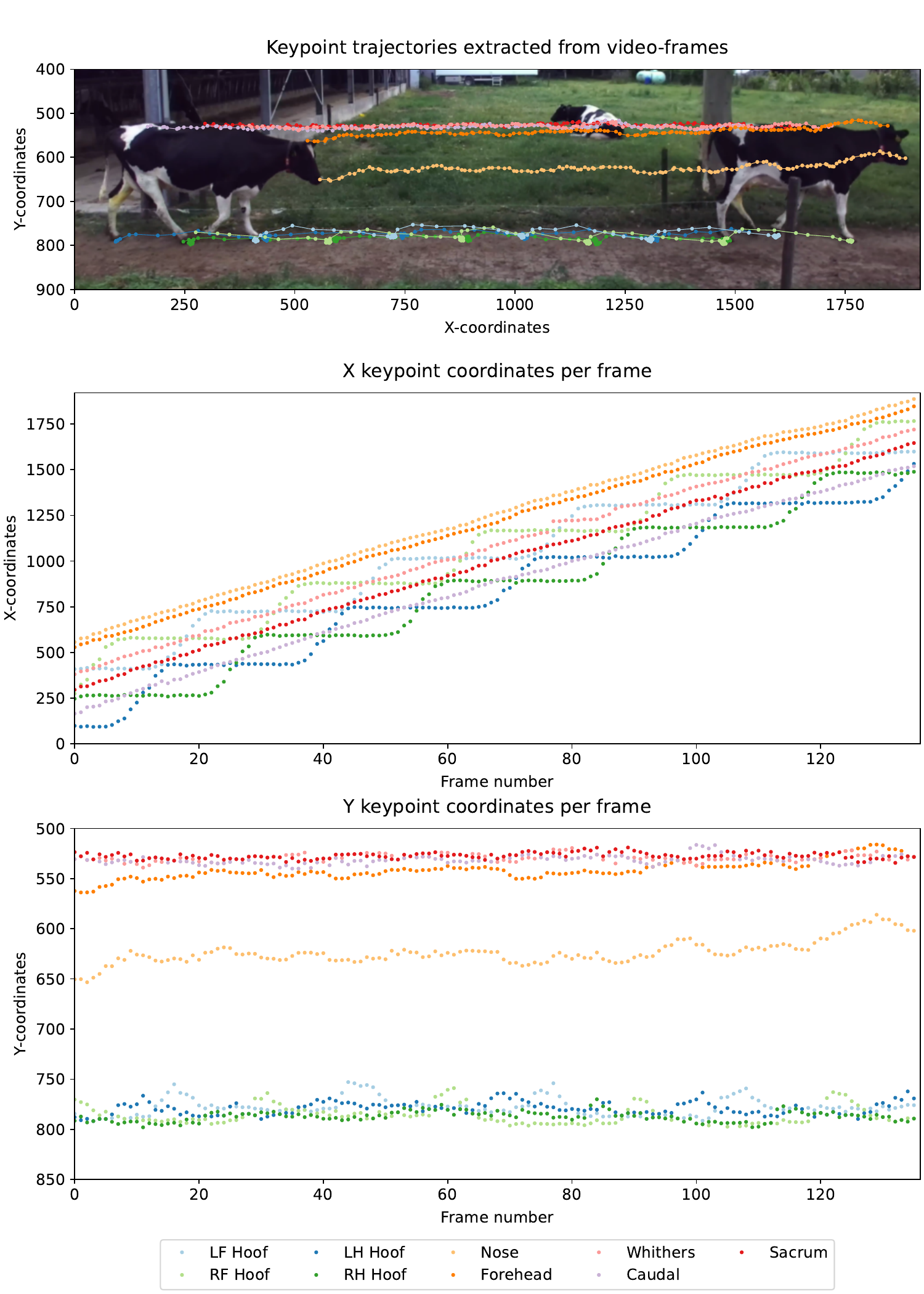}
    \caption{Example of the keypoint trajectories extracted with T-LEAP from a video of a cow presenting a \emph{normal} gait. 
    The top figure is augmented with the first and last frame of the video and shows the $(x,y)$ coordinates of each keypoint extracted from all the video frames. The middle and bottom figures show the keypoints' $x$ coordinates per frame, and $y$ coordinates per frame, respectively.}
    \label{fig:trajectory-example-normal}
\end{figure}

\begin{figure}[ht]
    \centering
    \includegraphics[width=0.8\textwidth]{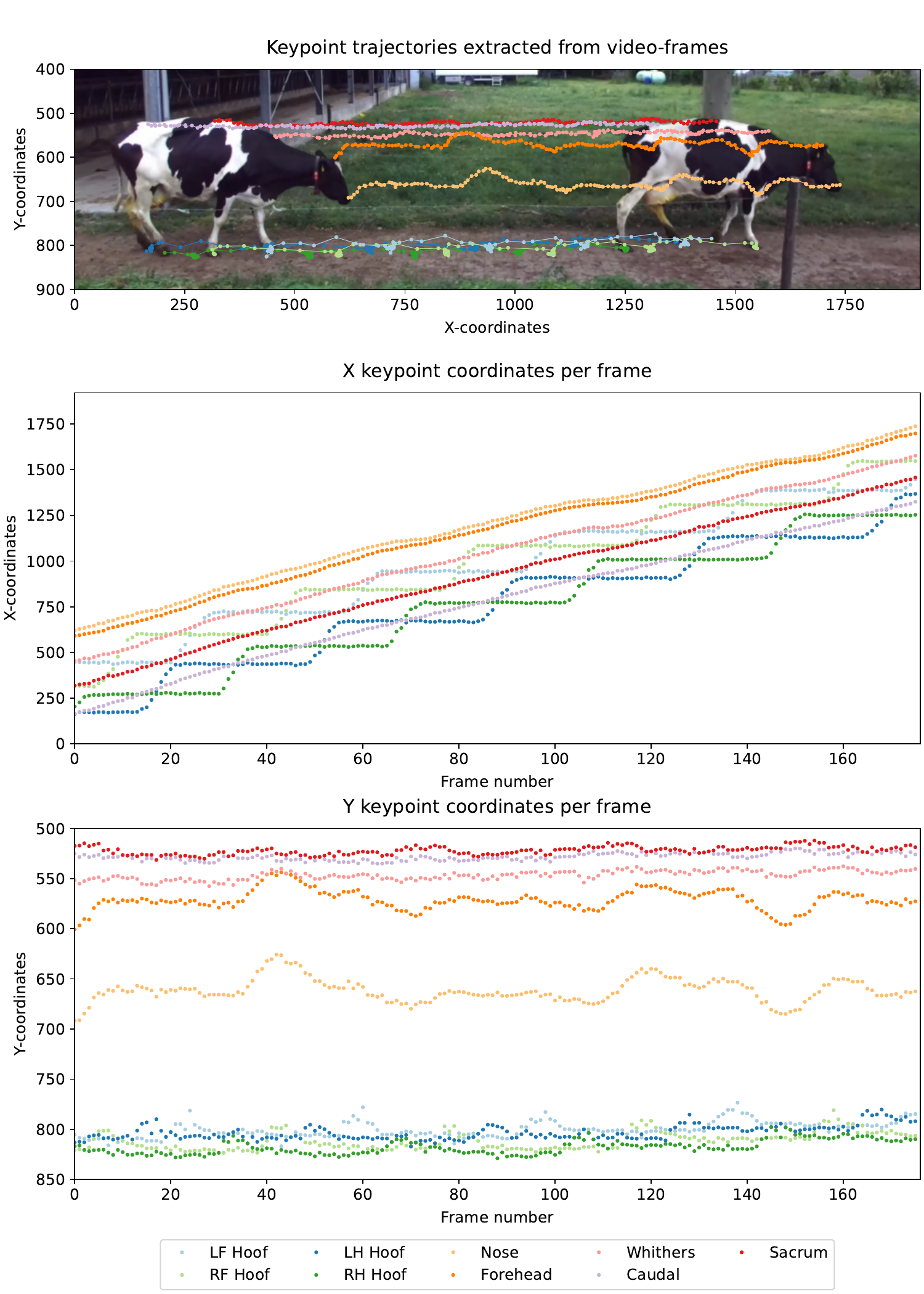}
    \caption{Example of the keypoint trajectories extracted with T-LEAP from a video of a cow presenting a severely \emph{lame} gait. 
    The top figure is augmented with the first and last frame of the video and shows the $(x,y)$ coordinates of each keypoint extracted from all the video frames. The middle and bottom figures show the keypoints' $x$ coordinates per frame, and $y$ coordinates per frame, respectively.}
    \label{fig:trajectory-example-lame}
\end{figure}

The keypoint trajectories were automatically extracted with the T-LEAP pose estimation model introduced in~\cite{russello_t-leap_2021}.
In short, T-LEAP is a pose estimation model that was trained on videos of walking cows to track nine anatomical landmarks (keypoints) of the cow's hooves, head, and back (Figure~\ref{fig:9keypoints}).
The keypoint trajectories, illustrated in Figures~\ref{fig:trajectory-example-normal} and~\ref{fig:trajectory-example-lame}, represent the motion in the 2D image-plane of the nine keypoints through each video frame.
Figure~\ref{fig:trajectory-example-normal} shows examples of keypoint trajectories for a normal gait, and Figure~\ref{fig:trajectory-example-lame} for a lame gait. 
For a more detailed description of the dataset, we refer the reader to~\cite{russello2024video}.

\FloatBarrier

\subsection{Data augmentation}\label{sssec:data-aug}
Data augmentation was performed on the \emph{training set} through temporal cropping and random jitter as described below.
For the \emph{test set}, no random jitter was applied and fixed sequences of 30, 60 and 90 frames long were extracted from the middle portion of the keypoint trajectories.

\paragraph{Random Temporal cropping}
Given that keypoint trajectories varied between 90 and 207 frames, while the experiments at hand required fixed sequence lengths of $T \in \{30, 60, 90\}$ frames, %
random cropping was applied to the data, allowing fixed sequence lengths, while augmenting the data.
That is, given a keypoint trajectory of $F$ frames, the starting frame was randomly selected from a discrete uniform distribution over $[0, F-T]$. This approach ensured that, for each training epoch, the model encountered different segments of the keypoint trajectories. 

\paragraph{Random jitter}
Similar to~\cite{karoui_deep_2021}, we augmented keypoint coordinates with random Gaussian noise to increase model robustness. 
Specifically, we applied jitter to each keypoint coordinate by randomly sampling from a normal distribution where the mean was set to the original coordinate value and the standard deviation was defined as one percent of the head length (measured as the Euclidean distance between head and nose keypoints).

\subsection{Lameness detection using BLSTMs}
In this study, we used a Bidirectional Long-Short-Term Memory (BLSTM) classification model~\citep{graves2005framewise} to detect lameness based on the keypoint trajectories of walking cows.
LSTMs~\citep{hochreiter1997long} are a type of Recurrent Neural Network (RNN) that excel at capturing long-term relationships in sequential data and learning complex temporal patterns. 
As their name suggests, RNNs use recurrent connections, that is, the output of a neuron at one time step loops back as an input to the neuron at the next time step. This allows to capture temporal dependencies and patterns within sequences.
An LSTM is a type of RNN designed to overcome the limitations of RNNs and allow to capture long-term dependencies in sequential data~\citep{hochreiter1997long}.
An LSTM unit is typically composed of a cell-state controlled through three gates: a forget gate, an input gate, and an output gate.
The forget gate controls what information to discard from the previous state, the input gate controls what new information to store in the cell-state , and the output gate controls what information of the cell-state to output.
At each time step, the LSTM takes the current input and previous hidden state, then uses these gates to selectively update its memory and produce an output. 
This gating mechanism allows LSTMs to learn long-term dependencies in sequences - they can remember important information from many time steps ago while forgetting irrelevant details~\citep{gers2002learning}. 
This makes them useful for analyzing gait patterns~\citep{lefebvre_blstm-rnn_2013, battistone_tglstm_2019}, since the relation between the sequential change in the pose of the animal and the presence of lameness is complex.

The main difference between unidirectional and bidirectional LSTMs lies in how they exploit the temporal information in input sequences. 
Unidirectional LSTMs process the input sequence in a single direction, either from the start to the end of the sequence (forward direction) or from the end to the start (backward direction).
A forward LSTM only includes information from the past to predict the current timestep, whereas a backward LSTM includes information from the future to predict the current timestep. 
As their name suggests, bidirectional LSTMs combine a forward LSTM and a backward LSTM, both of which are connected to the same output layer. 
This means that for any timestep in a sequence, the BLSTM utilizes sequential information from the points before and after that timestep~\citep{graves2005framewise}.
This is useful if future features provide different information to the current timestep than past features, which is expected with asymmetric signals, such as gait. 

\subsection{BLSTM model architecture}

\begin{figure}[ht]
    \centering
    \includegraphics[width=1\linewidth]{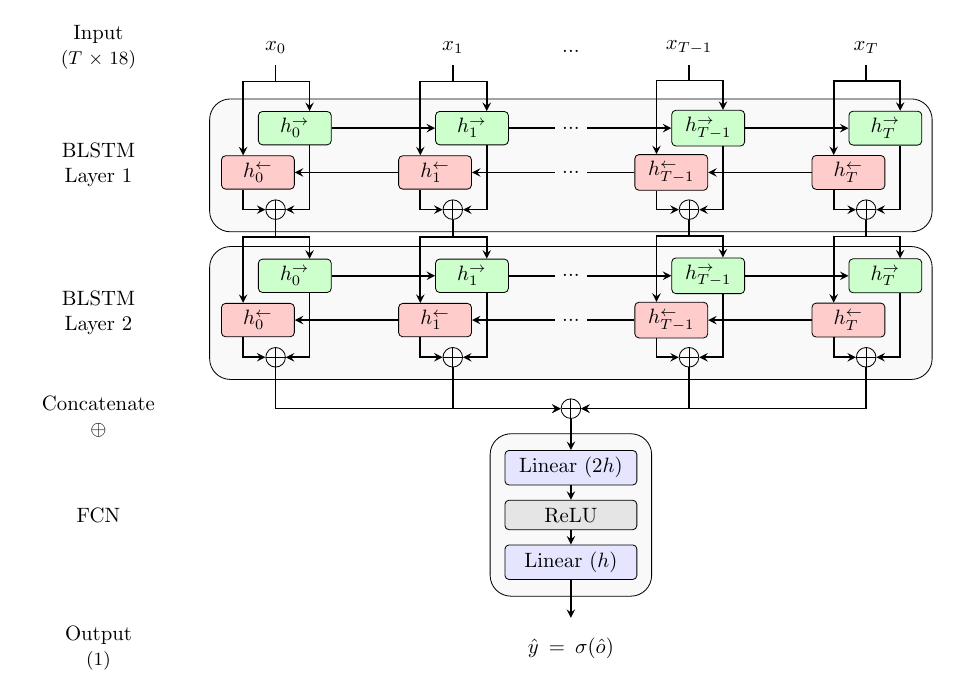}
    \caption{Architecture of the 2-layer BLSTM classifier, where $T$ corresponds to the sequence length (30, 60 or 90), $h$ corresponds to the number of hidden units (128 or 256). The green rectangles represent the forward LSTMs, and the red rectangles represent the backward LSTMs. Note that the 3-layer BLSTM architectures have an additional BLSTM layer, but their FCN remains the same.}
    \label{fig:lstm-arch}
\end{figure}

The proposed BLSTM classification model, illustrated in Figure~\ref{fig:lstm-arch}, receives sequences of keypoints as input and classifies the sequence as either~\emph{normal} or~\emph{lame}.
The model architecture consists of two parts: a BLSTM neural network followed by a Fully Connected Network (FCN) (Figure~\ref{fig:lstm-arch}).
 We empirically selected different BLSTM architectures, consisting of two or three layers of 128 or 256 hidden units ($h$). 
The BLSTM neural network presented in Figure~\ref{fig:lstm-arch} is composed of two BLSTM layers; our three-layers architecture has one additional BLSTM layer, but the FCN remains the same.

The input ($x$) consists of sequences of shape $T\times18$, where $T$ corresponds to the length of the sequence, and $18$ corresponds to the flattened vector containing the 2D image coordinates of $9$ keypoints. 
The input is passed to the BLSTM sequentially, i.e. from $x_0$ to $x_T$.
A BLSTM layer combines a forward and backward LSTM, each containing $h$ hidden units. 
The forward LSTM processes the input sequence in the forward direction (from the first item in the sequence, $x_0$, to the last item, $x_T$) and the backward LSTM processes the input sequence in the backward direction (from $x_T$ to $x_0$).
The outputs of the forward and backward LSTMs are concatenated and passed to the next BLSTM layer.
The outputs of the last BLSTM layer are concatenated and given to the classification layer (FCN), which is composed of two fully-connected layers of input size $2h$ and $h$.
The classification layer outputs a logit ($\hat{o}$), that is, an unnormalized prediction.
During inference, a sigmoid function ($\hat{y} = \sigma (\hat{o})$) is used to normalize the output  between 0 and 1, representing the probability of lameness. 
A cow is classified as lame if $\hat{y} \geq 0.5$.

\FloatBarrier

\subsection{Training procedure}

The models were implemented using the PyTorch deep-learning framework (version 2.1.0)~\citep{ansel2024pytorch}.
The experiments were run on a computer equipped with an AMD Ryzen 9 5900X CPU, an NVIDIA GeForce RTX 4070 Ti-SUPER GPU, and 64 GB of RAM.

Given the limited size of the dataset and to increase the reliability of the results, we followed the same procedure as~\cite{russello2024video}, and split the dataset into training and validation sets with a stratified 5-fold cross-validation with grouping. 
The folds were grouped on the ID of the cows, ensuring that individual cows appeared in the training or validation set, not both.
To ensure an equal distribution of the classes in each training fold, we used PyTorch's weighted random sampler~\citep{ansel2024pytorch} to address class imbalance through minority class oversampling.
All models were trained with 5-fold cross-validation with batches of size 8 for 100 epochs. 
We used the AdamW-AMSGrad optimizer~\citep{reddi2019convergence, loshchilov2017decoupled} with a learning rate scheduler that reduced the learning rate by a factor of two every 50 epochs. 
Gradient clipping~\citep{pascanu2013difficulty} with a threshold of $0.5$ was applied to prevent vanishing and exploding gradients, a common problem with LSTMs~\citep{bengio1994learning}.
The learning rate, weight decay, and dropout hyperparameters were tuned using a flat cross-validation approach.
That is, the hyperparameters were first optimized per fold using a grid search over a 5-folds cross-validation, and the models were then re-trained on the same 5-folds cross-validation with the best set of hyper-parameters.
We used flat cross-validation as it is computationally less expensive than nested cross-validation and generally results in the selection of an algorithm of similar quality to that selected via nested cross-validation~\citep{wainer2021nested}.

\subsection{Evaluation metrics}

The performance of the models was evaluated using the following metrics: accuracy, macro F1-score, sensitivity, and specificity. 
The F1-score was macro-averaged; that is, the metric was calculated per class (\textit{normal}/\textit{lame}) and then averaged. 
The evaluation metrics were averaged over the five cross-validation folds.
The differences between the different models were statistically tested using the McNemar's test with binomial distribution~\citep{mcnemar1947note}, setting $p < 0.05$ as the level of significance.

\subsection{Experiments}

\subsubsection{Different model architectures}

We first explored different network architectures, with two and three BLSTM layers of 128 and 256 hidden units, later referred to  as $2 \times 128$,  $3 \times 128$,  $2 \times 256$, and $3 \times 256$.
Although these architectures were chosen empirically, they were inspired from~\cite{graves2013speech}, who proposed, among others, BLSTM architectures for speech recognition consisting of 2 and 3 layers of 250 hidden units.
These first experiments were conducted with sequences of $T=90$ frames, since this was the length of the shortest video in the dataset.

\subsubsection{Comparison with reference model}

This study sought to compare the performance in detecting lameness through learned locomotion features (proposed model) to previously established manually-designed locomotion traits (reference model).
We then performed a direct comparison with~\cite{russello2024video}, and compared our results to their best model, a Support Vector Machine (SVM) classifier trained with six locomotion features on the same dataset.

Briefly, \cite{russello2024video} used the same dataset of keypoint trajectories and applied outlier detection and filtering to correct and smoothen noisy keypoint trajectories. 
These keypoint trajectories were then used to compute six locomotion traits, namely back-posture, head-bobbing, tracking distance, stride length, stance duration, and swing duration. 
The locomotion traits were then used as input features to train multiple classifiers, such as decision trees or SVMs, to detect lameness.

Unlike in~\cite{russello2024video}, no filtering was applied to the BLSTM input data, since BLSTMs can typically handle some level of noise.
To compare the impact of filtering, we trained the reference SVM classifier with locomotion traits computed from the filtered keypoint trajectories, as well as from the raw, unfiltered keypoint trajectories.

\subsubsection{Different sequence lengths}
The sequence duration is an important factor to take into account when detecting lameness, since healthy cows tend to walk faster (1.3 seconds per stride) than lame cows (1.5 seconds per stride)~\citep{flower2005hoof}.

The first set of experiments were conducted with sequences of 90 frames (three seconds of video data), since this was the length of the shortest video in the dataset.
We further extended our analysis by evaluating our models on shorter sequences, namely sequences lengths of 30 and 60 frames, which corresponds to one and two seconds of video data, respectively.
To this end, we re-trained and evaluated the best-performing BLSTM architecture on sequences of 30 and 60 frames.

\section{Results}

The results of the different BLSTM architectures and sequence lengths, compared to the SVM approach on the same dataset, are presented and discussed in the following subsections.

\subsection{Different BLSTM architectures}

The upper part of Table~\ref{tab:res-arch} presents the results of various BLSTM architectures using 90-frame sequences.
The accuracy ranged from 83\% to 85\% and macro F1-scores between 83\% and 84\%. The sensitivity ranged from 78\% to 83\%, and the specificity ranged from 85\% to 89\%. 
The BLSTM $2\times256$ and BLSTM $3\times128$ architectures performed the best overall, and had an F1 score of 84\%.
The BLSTM $2\times256$ had a higher accuracy, while the BLSTM $3\times128$ had a higher sensitivity and specificity.
The BLSTM $2\times128$ and BLSTM $3\times256$ architectures performed slightly worse than the other two. 
In general, there was minimal variation in performance between the different architectures, and only the BLSTM $2\times128$ was significantly different from the other architectures.

\begin{table}[h]
    \centering
    \caption{Evaluation results of the different BLSTM architectures for sequence lengths of 90 frames, followed by the results of the SVM approach of~\cite{russello2024video} on the same dataset.
    The results are presented in percents (\%).
    The best results are highlighted in \textbf{bold} and the second best results are \underline{underlined}.}
    \label{tab:res-arch}
    \begin{adjustbox}{width=\columnwidth,center}
        \begin{tabular}{@{}llllll@{}}
        \toprule
        Model               & N\# frames     & Accuracy     & F1 (macro)     & Sensitivity    & Specificity       \\ \midrule
        
        BLSTM $2\times128$   & 90             & 83.48        & 82.63        &   78.48        &   \textbf{88.74}              \\
        BLSTM $3\times128$   & 90             & \underline{84.47}        & \textbf{83.70}        &    \textbf{82.74}               &    \underline{87.78}             \\
        BLSTM $2\times256$   & 90             & \textbf{84.59}        & \textbf{83.70}       &     \underline{81.94}              &  87.07 \\
        BLSTM $3\times256$   & 90             & 83.11        & 82.37      &      81.77             &    85.07       \\ 
        \midrule
        Gait features + SVM (filtering)  & 124 ($\pm$ 24) & 80.07       & 78.70       & 76.78   &  81.15 \\ %
        Gait features + SVM (no-filtering)    & 124 ($\pm$ 24)                & 75.17           & 73.38        & 70.70        &  76.26 \\ %
        \bottomrule
        \end{tabular}
    \end{adjustbox}
\end{table}

\subsubsection{Comparison with reference model}

The lower part of Table~\ref{tab:res-arch}, presents the results of~\cite{russello2024video} for the same dataset.
The evaluation results are displayed for their best model, that is, a radial-kernel SVM classifier, with and without keypoint filtering.

The performance of the SVM classifiers were lower across all metrics, for both approaches with and without keypoint filtering. 
For instance, without filtering on the keypoint trajectories, the F1-score was about nine percentage points lower than the BLSTM.
When applying filtering, the performance of the SVM classifier increased dramatically, but remained significantly lower than that of the BLSTM with a F1-score lower by five percentage points. 
The predictions from the BLSTM models were all significantly different from the predictions of the SVM models.

\subsection{Different sequence lengths}

Table~\ref{tab:res-seq} presents the results of the BLSTM $3\times128$ model trained with sequences of 90, 60, and 30 frames.
Using 90-frame sequences led to the best overall performance.
Although the performance was lower accross all metrics with sequences of 60 and 30 frames, the difference was not statistically significant.

\begin{table}[h]
    \centering
    \caption{Evaluation results of the BLSTM model trained with different sequence lengths.
    The results are presented in percents (\%).
    The best results are highlighted in \textbf{bold}.}
    \label{tab:res-seq}
    \begin{adjustbox}{width=\columnwidth,center}
        \begin{tabular}{@{}llllll@{}}
        \toprule
        Model               & N\# frames     & Accuracy     & F1 (macro)     & Sensitivity    & Specificity       \\ \midrule
                           & 90             & \textbf{84.47}        & \textbf{83.70}        &    \textbf{82.74}               &    \textbf{87.78}             \\
        BLSTM $3\times128$   & 60             & 83.08        & 82.23        &    82.60               &    83.73             \\
                            & 30             & 83.07        & 82.23        &    82.07               &    83.91             \\

        \bottomrule
        \end{tabular}
    \end{adjustbox}
\end{table}

\section{Discussion}

\subsection{Lameness detection}\label{ssec:disc-lameness-detection}

The primary objective of this study was to investigate whether a learning-based end-to-end approach could improve lameness detection over conventional feature-based methods in recent literature~\citep{russello2024video}. 
We conducted a direct comparison between our BLSTM approach and a feature-based SVM approach \cite{russello2024video}, using identical experimental conditions.
That is, we used the the same dataset of 272 cow videos, with the same labels, and same keypoint trajectories generated by the T-LEAP pose estimation model.

All proposed BLSTM architectures substantially outperformed the SVM model, with BLSTM $3\times128$ and $2\times256$ achieving the highest accuracy. 
These results suggest that learning temporal gait patterns directly through BLSTMs offers improved performance compared to manual feature engineering approaches. 
A key finding emerged regarding robustness to noise: the SVM model required a prepossessing step to correct outliers and filter keypoint trajectories and its accuracy declined by 5 percentage points when using unfiltered data.
In contrast, our BLSTM models achieved better results even when trained on unfiltered trajectories.
This robustness to noisy keypoint detections suggests that BLSTMs would be more resilient to real-world challenges such as occlusions or corrupted video frames. 
Further research could explore the extent of this robustness by systematically evaluating BLSTM performance under different noise levels in the validation data. 
Such analysis could help establish minimum accuracy requirements for pose estimation models in lameness detection applications.

We approached lameness detection as a binary classification task, being either "\textit{normal}" or "\textit{lame}" walking, rather than multi-class locomotion scoring. 
The decision to merge the~\cite{sprecher_lameness_1997} 5-level locomotion scale (normal (1), mildly lame (2), moderately lame (3), lame (4), severely lame (5)) into a binary classification (normal (1) vs. lame (2-5)) was motivated in~\cite{russello2024video} as a way to address the dataset limitations: its relatively small size and skewed distribution toward lower lameness levels. 
As discussed in~\cite{russello2024video}, finer locomotion scoring would likely require a larger dataset with better representation of severe lameness cases.
Furthermore, the performance of the classifiers may be limited by the quality of the ground-truth data, as some bias from the subjective nature of locomotion scoring likely persisted even after data curation. 

The T-LEAP pose estimation model was originally trained on 17 keypoints~\citep{russello_t-leap_2021}, which included two more leg keypoints at the fetlock and carpal/tarsal joints. 
Here, we used nine of these keypoints (four hooves, three back keypoints, forehead, and nose) to represent the gait kinematics.
We used this subset of keypoints to allow a direct comparison between our BLSTM approach and the SVM approach in~\cite{russello2024video}, who used these nine keypoints to compute hand-crafted locomotion features.
Since BLSTMs can capture complex temporal patterns, incorporating additional keypoints could allow richer gait kinematics representations.
For instance, including keypoints at the fetlock and carpal/tarsal joints would include information of the touch and release angles, as well as joint angle velocities, which may vary with lameness severity~\citep{pluk_automatic_2012}. 
Investigating the benefits of additional keypoints for lameness detection is left to future work.

The dataset used by~\cite{russello2024video} contained videos that were selected when cows walked continuously, without distraction or interruption. 
As a result, the models were trained and evaluated under the assumption of uninterrupted locomotion. 
However, the impact of cows stopping on lameness detection performance remains unexplored. 
This potential selection bias might be a limitation, as real-world scenarios frequently involve cows pausing their locomotion. 
Future research should evaluate the degradation in model performance when cows exhibit natural behaviors.
This would provide insights into the applicability of these detection systems in farm environments.

\subsection{Different sequence lengths}

The impact of sequence length on lameness detection was evaluated on the $3\times128$ BLSTM model, testing sequences of 30, 60, and 90 frames (corresponding to 1, 2, and 3 seconds of video at 30 FPS). 
While 90-frame sequences achieved the best performance, the results from shorter sequences were comparable, suggesting that lameness detection might be feasible with only one second of video data.

Looking at the stride duration in our dataset, healthy cows required 31 frames on average (1.03 seconds) to complete a stride (hoof strike to hoof strike), and severely lame cows requiring up to 45 frames (1.5 seconds). 
Although 30-frame sequences might not capture a complete gait cycle (i.e., one stride per leg), they appear to provide sufficient information for accurate detection. 
This can be attributed to several factors. 
First, even within a 30-frame window, the model can assess walking speed, which decreases with moderate to severe lameness~\citep{blackie_associations_2013}.
Second, with slower, lame cows, these 30-frames sequences capture fewer strides, and while they might miss the stride of the affected leg, they might include compensatory movements in non-lame limbs that cows use to alleviate pain~\citep{blackie_associations_2013}.

Despite the promising results with shorter sequences, we recommend using longer sequences when possible.
Shorter sequences risk missing critical gait anomalies, particularly given that cows, being stoic animals, often attempt to hide their pain. 
Longer sequences provide more opportunities to identify subtle anomalies in gait patterns and establish a more comprehensive assessment of the animal's locomotion. 

Further research in understanding the minimum required sequence length could allow for an optimal camera placement in farms.
For instance, in space-limited or cluttered environments, the animals might only be visible for short periods of time.

\subsection{Comparison with related work}

Performing direct comparisons between our approach and related work presents significant challenges due to variations in data, methodologies, and evaluation metrics. 
In subsection~\ref{ssec:disc-lameness-detection}, we conducted a direct comparison with~\cite{russello2024video} by using the same dataset and leveraging their open-source code, which enabled us to demonstrate our method's improvements objectively.

In the following paragraphs, we discuss our approach in relation to other directly relevant studies.
However, these comparisons are constrained by the absence of publicly available data and code from these works. 
Without reproducing their experiments (which would be prohibitively time-consuming), direct one-to-one comparisons remain infeasible. Nevertheless, we compare our results and contrast our findings within these limitations. 
To advance reproducibility and promote open-source research practices, we commit to making our data and code publicly available upon publication and encourage similar transparency in future research.

\cite{karoui_deep_2021} developed a lameness detection system based on keypoint trajectories obtained from physical reflective markers placed on the cows legs. 
They extracted four keypoint trajectories per leg, and generated additional synthetic data by adding random noise to the keypoint trajectories, resulting in a dataset of 24000 samples.
The keypoint trajectories were used to train a LeNet CNN \cite{lecun1998gradient} for binary classification, achieving both accuracy and F1-score of 91\%.
Our approach differs in that rather than using physical markers, we used a markerless pose estimation model, which is less invasive and easier to deploy in practice. 
Furthermore, physical markers are more prone to skin-movement errors in kinematic data due to skin displacement while walking~\citep{bergh2014skin}.
In addition to the leg keypoints, we also included keypoints on the back and head regions, allowing us to cover a broader range of gait patterns.
Their experiments showed that generating new data by adding 5\% noise variation yielded the best results, and while we similarly augmented our data using Gaussian noise with 1\% standard deviation, we did not extensively study its impact. 
However, based on Karoui et al.'s findings, we believe this augmentation technique contributed to improving our models' robustness and generalization capabilities.

\cite{wu_lameness_2020} developed a lameness detection system using YOLOv3 for leg localization in videos of walking cows. 
Their approach extracted the relative step sizes by calculating the horizontal distance between left and right legs in each frame. 
These sequences of step-size vectors were used as gait-features for various classification models.
Their evaluation on 700 videos using 10-fold cross-validation achieved high accuracy metrics across different architectures: Decision Trees (90.5\%), Support Vector Machines (94.3\%), LSTM (98.6\%), and BLSTM (99\%). 
The performance gap between their SVM and BLSTM (4-5 percentage points) aligns with our findings. 
While their overall accuracy exceeded ours, this can be attributed to their larger training dataset: their 10-fold cross-validation used 630 videos for training, while our 5-fold cross-validation used, on average, 228 videos for training. 
When they reduced the training set to 350 videos using 2-fold cross-validation, performance decreased substantially (SVM: 82.9\%, BLSTM: 86.7\%).
Furthermore, despite using significantly shorter sequences (30-90 frames versus their 500 frames), our system achieved similar performance.
Our approach differs in the following aspects. 
While \cite{wu_lameness_2020} focused solely on the horizontal movements of the legs, we included both horizontal and vertical motion of keypoints from the legs, back, and head, providing richer spatial information.
Furthermore, they first computed step-size vectors from the leg coordinates, and used this gait-feature as an input to their models, whereas we used the raw keypoint trajectories, thus eliminating the need for manual feature engineering and pre-processing.

\cite{jiang_single-stream_2020} proposed a lameness classification approach combining optical-flow maps and a BLSTM classifier.
Their study used a dataset of 1080 videos of walking cows (756 for training, 324 for testing) of 125 to 1000 frames long, with lameness severity scored on a 4-point scale (normal, slight, moderate, and severe).
Sequences of optical-flow maps were generated from the videos frames, and a DenseNet CNN~\citep{huang2017densely} was used to extract spatial features from each time-step of the optical-flow maps. 
A BLSTM classifier was then trained to classify 4-levels of lameness from sequences of spatial features, achieving a classification accuracy of 95\%. 
Our approach differs primarily in how spatial data is extracted from video frames. 
We focused on tracking nine specific keypoint coordinates, while their method used optical-flow maps to capture motion information from the entire body. 
Although their approach provided richer spatio-temporal information, it resulted in significantly higher input dimensionality to the BLSTM ($T \times 224 \times 224$ compared to our $T \times 18$). 
This increased complexity made their learning process more computationally demanding, requiring pre-training and a larger training dataset.
Additionally, they studied the impact of sequence lengths on the classification accuracy, and had an improvement from 85\% using 30-frame sequences to 95\% using 60-frame sequences. 
Notably, they showed that longer sequences particularly benefited the detection of moderate and severe lameness cases, which can be attributed to the slower walking pace of lame cows.
Our findings differ from theirs in that regard, as we observed minimal accuracy differences between the BLSTM models using 30, 60, and 90 frames. 
These contrastive findings may be attributed to differences in dataset composition.
First, they performed classification across four, evenly balanced, lameness severity levels, whereas we performed a binary classification.
Even though our dataset with binary labels was fairly balanced, the binary labels were merged from a five-point lameness scale, with skewed distribution towards non- and slightly-lame cows and with fewer severe cases. 
Merging the lameness levels decreased the granularity in the data, thereby potentially masking the benefits of longer sequences for detecting severe cases.

To summarize, this subsection compared our pose-estimation and BLSTM approach against three directly relevant studies: \cite{karoui_deep_2021} used physical reflective markers on cow legs with LeNet CNN achieving 91\% accuracy, but required invasive marker placement; 
\cite{wu_lameness_2020} employed YOLOv3 for leg localization and achieved up to 99\% accuracy with BLSTM on 700 videos, though performance dropped significantly with smaller training sets; 
and \cite{jiang_single-stream_2020} combined optical-flow maps with BLSTM for 4-level lameness classification, achieving 95\% accuracy but requiring higher computational complexity and larger datasets.

\section{Conclusion}

In this paper, we developed a lameness-detection approach combining pose estimation and BLSTMs. 
Motion sequences of nine keypoints (located on the cows' hooves, head and back) were extracted from videos of walking cows with the T-LEAP pose estimation model.
The trajectories of the keypoints were then used as an input to a BLSTM classifier that was trained to perform binary lameness classification. 
The experiments consisted of comparing our proposed four BLSTM architectures to an established method relying on manually-designed locomotion traits, as well as comparing the performance across short sequence lengths (1,2 and 3 seconds of video data).

Across the proposed model architectures, the BLSTM $3\times128$ and $2\times256$ lead to the overall best performance, with an accuracy of 84.5\%, against 80.1\% accuracy for the feature-based approach. 
Furthermore, our pose-estimation and BLSTM approach achieved comparable performance to other studies that used BLSTM lameness classifiers~\citep{wu_lameness_2020, jiang_single-stream_2020}. 
BLSTM classifiers demonstrated similar performances across different sequence lengths.
Our model achieved an accuracy of 84.5\% when using 3-seconds sequences, and 83.1\% when using both 2-seconds and 1-seconds sequences, suggesting that lameness can be detected with as little as one second of video data.

Combining markerless pose-estimation to extract the movement of multiple keypoints with BLSTMs to learn temporal locomotion features offered several advantages. 
Using raw keypoint trajectories eliminated the need for manual feature engineering, while including leg, back and head keypoints provided rich spatial information. 
Furthermore, our approach was more efficient than previous studies that used BLSTM lameness classifiers~\citep{wu_lameness_2020, jiang_single-stream_2020}, since we worked with with significantly shorter sequences and smaller training datasets.

While our approach offers a promising direction in automatic lameness detection from videos, further research directions include: (1) systematically evaluating BLSTM robustness under varying noise levels to establish minimum accuracy requirements for pose estimation models, (2) investigating the benefits of additional keypoints for improved detection, (3) exploring fine-grained lameness classification on a multi-point scale rather than binary classification, 
(4) evaluating the performance of lameness detection in "real-world" conditions, and (5) establishing the minimum duration requirements of video clips.

\section*{Acknowledgments}
This publication is part of the project Deep Learning for Human and Animal Health (with project number EDL P16-25-P5) of the research program Efficient Deep Learning (\url{https://efficientdeeplearning.nl}) which is (partly) financed by the Dutch Research Council (NWO).

\section*{Declaration of interests}
The authors declare that they have no known competing financial interests or personal relationships that could have appeared to influence the work reported in this paper.

\section*{Declaration of generative AI and AI-assisted technologies in the writing process}
During the preparation of this work the authors used Claude (Sonnet 4) by Anthropic in order to refine selected sentences for clarity.
After using this tool, the authors reviewed and edited the content as needed and take full responsibility for the content of the published article.

\clearpage
\bibliography{gait-analysis-paper.bib}  %

\end{document}